\documentclass[runningheads]{llncs}
\usepackage{graphicx}
\usepackage{tikz}
\usepackage{comment}
\usepackage{amsmath,amssymb} % define this before the line numbering.
\usepackage{color}
\usepackage{booktabs}
\usepackage{multirow}
\usepackage{diagbox}
\usepackage{dblfloatfix} 
\usepackage{hyperref} 
\hypersetup{
hidelinks,
colorlinks=true,
linkcolor=black,
citecolor=black
}

\usepackage[accsupp]{axessibility}  % Improves PDF readability for those with disabilities.

\usepackage{alphabeta}
\usepackage{bm}

\begin{document}
% \linenumbers
\pagestyle{headings}
\mainmatter

\title{Lane Change Classification and Prediction with Action Recognition Networks} % Replace with your title
     
\begin{comment}
% \titlerunning{ECCV-22 submission ID \ECCVSubNumber} 
% \authorrunning{ECCV-22 submission ID \ECCVSubNumber} 
% \author{Anonymous ECCV submission}
% \institute{Paper ID \ECCVSubNumber}
\end{comment}
%******************

% \begin{comment}
\titlerunning{Lane Change Classification and Prediction}
\author{Kai Liang \and Jun Wang \and Abhir Bhalerao}
%% First names are abbreviated in the running head.
\authorrunning{Liang et al.}
\institute{Department of Computer Science, University of Warwick, UK\\
\email{\{kai.liang, jun.wang.3, abhir.bhalerao\}@warwick.ac.uk}}

% \end{comment}
%******************
\maketitle
%%%%%%%%%%%%%%%%%%%%%%%%%%%%%%%%%%%%%%%%%%%%%%%%%%%%%%%%%%%%%%%%%%%%%%%%%%%%%%%%%%%%%%%%%%%%%%%
\begin{abstract}
Anticipating lane change intentions of surrounding vehicles is crucial for efficient and safe driving decision making in an autonomous driving system. Previous works often adopt physical variables such as driving speed, acceleration and so forth for lane change classification. However, physical variables do not contain semantic information. Although 3D CNNs have been developing rapidly, the number of methods utilising action recognition models and appearance feature for lane change recognition is low, and they all require additional information to pre-process data. In this work, we propose an end-to-end framework including two action recognition methods for lane change recognition, using video data collected by cameras. Our method achieves the best lane change classification results using only the RGB video data of the PREVENTION dataset. Class activation maps demonstrate that action recognition models can efficiently extract lane change motions. A method to better extract motion clues is also proposed in this paper.~\footnote{The code is publicly available at~\href{https://github.com/kailliang/Lane-Change-Classification-and-Prediction-with-Action-Recognition-Networks}{https://github.com/kailliang/Lane-Change-Classification-and-Prediction-with-Action-Recognition-Networks}}

\end{abstract}

\section{Introduction}

Driving is one of the most common and necessary activities in daily life. It can however be  often seen as a boring and inefficient experience because of traffic congestion and the tedium of long journeys. In addition, finding a parking area in a city centre can be difficult. As a newly innovative technique, autonomous driving has the potential to alleviate these problems by taking place alongside manually driven vehicles. For instance,  shared autonomous vehicles are more economical than most modes of transportation, e.g., taxi and ride-hailing services, and require fewer parking spaces \cite{92}. Furthermore, autonomous vehicles can dwindle the number of traffic accidents caused by infelicitous human behaviours or subjective factors, e.g., fatigue and drunk driving \cite{93}.
Autonomous vehicles can also operate in a coordinated mode by sharing their trajectories and behavioural intentions to improve energy efficiency and safety \cite{25}. Consequently, autonomous vehicles are slowly becoming a new part of the infrastructure in cities \cite{92}. 

\begin{figure}[t]
    \centering
    \includegraphics[width= 0.82\textwidth]{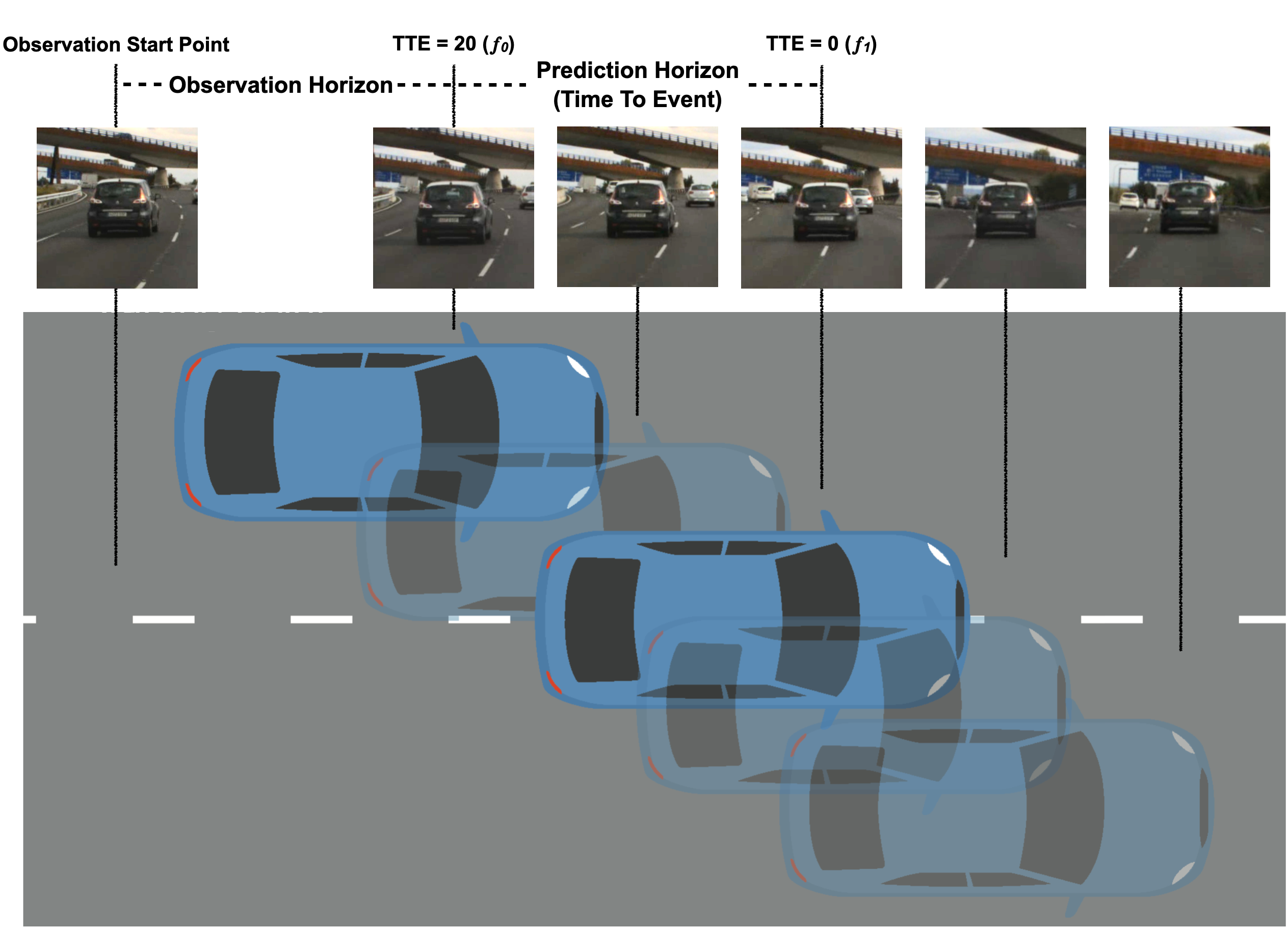}
    \caption{A lane change event where a  vehicle performs a right lane change. $f_0$ denotes the frame at which lane change starts and $f_1$ is the frame at which the rear middle part of the target vehicle is just between the lanes. The Observation Horizon is defined as 40 frames (4 seconds at 10 FPS before $f_0$). The Prediction horizon or Time To Event (TTE) (on average of length 20 frames at 10 FPS) is defined as the time from $f_0$ to $f_1$. }
    \label{fig:illst_tte}
\end{figure}

Developing an applicable driving system is faced however with huge difficulties, since it not only requires recognizing object instances in the road, but also requires understanding and responding to a complicated road environment, such as the behaviours of other vehicles and pedestrians, to ensure safety and efficiency. Lane change recognition, aiming to anticipate whether a target vehicle performs left lane change, right lane change or remains in its lane, plays an important role in the autonomous driving system. Autonomous vehicles and human-driven vehicles will need to coexist on the road for some time to come \cite{33}, thus  lane change prediction is a tricky task since it has to cope with the complicated road environment shared with human drivers. For instance, in highway scenarios, surrounding vehicles may perform cut-in and cut-out manoeuvres unexpectedly, sometimes at high speeds. In a cut-in lane change scenario, a vehicle from one of the adjacent lanes merges into the lane right in front of the ego vehicle. A cut-out lane change scenario typically happens when the vehicle leaves its current lane to avoid slower vehicles ahead. Autonomous vehicles need to respond safely to these actions of human-driven vehicles while planning their trajectories. Hence, predicting and understanding the possible lane changes should be done at the {\em earliest} moment. Figure \ref{fig:illst_tte} illustrates an example of a right lane change where the target vehicle has its manoeuvre at $ f_0 $, the start of the Prediction Horizon, but is observed for several seconds before this point. The challenge then is to correctly predict the lane change event as early as possible. 

Most of current methods~\cite{30,32,34,35,36,37} use physical variables, e.g., driving speed, acceleration, time-gap, heading angle, yaw angle, distances, etc. for lane change recognition. Nevertheless, physical variables cannot represent the type of target objects as they do not contain enough semantic information, whereas knowing the type of road agents can help autonomous vehicles make decisions accordingly. Experienced drivers predict the potential lane changes of surrounding vehicles via pure visual clues, and adjust their speed accordingly, or even change lane if necessary. This ability enables them to anticipate potentially dangerous situations and respond to the environment appropriately.  Recently, action recognition models in computer vision~\cite{4,5,6,7,8,17,18} have demonstrated improvements on human activity recognition. 3D action recognition models efficiently extract spatial and temporal clues by only using visual information such as video data. 
A few works utilising action recognition methods for intelligent vehicle systems have been proposed. 
Nonetheless, these methods all require additional annotations to pre-process data, e.g., identifying a target vehicle in the video stream~\cite{2,25,44}. 

To address this problem, we propose a novel end-to-end framework involving two approaches for lane change recognition. Specifically, our method takes advantage of the recent powerful action recognition models and mimics a human drivers' behaviours by only utilizing the visual information. The first approach (RGB+3DN) only utilises the similar visual information that humans use to recognise actions, i.e. raw video data collected by cameras. This method is tested by 3D action recognition networks including I3D networks \cite{4}, SlowFast networks \cite{7}, X3D networks \cite{8} and their variants. The second approach (RGB+BB+3DN) employs the same action recognition models, but uses information of vehicle bounding boxes to improve the detection performance. The vehicle bounding box information is embedded into each frame of the RGB video data to enhance the detection and prediction and passed to a pre-trained action recognition model. 
We further study the spatial and temporal attention region of activated by the recognition models by generating class activation map \cite{98CAM}. Furthermore, we propose a better way to extract motion clues by optimizing the pooling stride.

Our main contributions can be summarized as follows:
\begin{enumerate}
\item We introduce an end-to-end framework for lane change recognition from front-facing cameras involving two action-recognition approaches. 
\item We perform an extensive set of experiments and achieve state-of-the-art lane classification results using raw RGB video data. 
\item We generate Class Activation Maps (CAMs) to investigate the temporal and spatial attention region of the trained 3D models.

\item We propose to utilize a smaller temporal kernel size and demonstrate it can better extract motion clues.

\item We compare the performance of seven CNN action recognition networks to perform lane change classification and prediction of surrounding vehicles in highway scenarios.
\end{enumerate}

Following an introduction, Section 2 gives an in-depth review previous works for lane change recognition and the recent development of action recognition methods, such as SlowFast networks~\cite{7}. In Section 3, an overview of the problem formulation is presented. Section 4 describes in detail the implementation  all the proposed recognition approaches. In Section 5, the performance and the evaluation metrics of the different methods is assessed. We conclude with a summary of our finding and make proposals for further work.

\section{Related Work}
\subsubsection{Lane Change Recognition with Physical Variables}

To handle lane change recognition problems, many existing works~\cite{32,34,35,36,37,43} normally employ physical variables to represent the relative dynamics of a target vehicle with its surrounding vehicles, e.g., driving speed, acceleration, time-gap, heading angle, yaw angle, distances, etc. Lane change events are predicted by analysing physical variables with classical Machine Learning methods~\cite{26,28,29,30,33,43}. Environment information is also introduced in some works ~\cite{32,95,96,97}. For example, Bahram et al.~\cite{32} utilise road-level features such as the curvature, speed limit and distance to the next highway junction. Lane-level features such as type of lane marking, the distance to lane end and number of lanes are also employed in their work. 

\subsubsection{Action Recognition for Lane Change Classification}
Owing to the improvements in computational hardware, CNN based prediction algorithms have been adopted to understand images and videos, and even obtain superior performance than humans in some fields such as image recognition, especially for fine-grained visual categorization. To this end, some research explores the potential of the CNN for lane change detection and classification. Autonomous vehicles can make use of these algorithms to anticipate situations around them \cite{44}. In 2014, Karpathy et al.~\cite{1} proposed the first Deep Learning action recognition framework, a single stream architecture that takes video clips as input. Simonyan et al.~\cite{2} proposed a two-stream based approach called C2D where a spatial stream takes still frames as input and a temporal stream takes multi-frame dense optical flow as input. Later, 3D convolution for action recognition was introduced by Tran et al. in 2015 \cite{3}, and since, 3D convolution methods have dominated the field of action recognition because they can efficiently extract spatial and temporal features simultaneously. Notably, the following sate-of-the-art action recognition models such as R3D \cite{5}, I3D \cite{4}, S3D \cite{17}, R(2+1)D \cite{6}, P3D \cite{18}, SlowFast \cite{7} and X3D \cite{8} employ 3D convolutions. 

I3D \cite{4} is a widely adopted 3D action recognition network built on a 2D backbone architecture (e.g., ResNet, Inception), expanding all the filters and pooling kernels of a 2D image classification architecture, giving them an additional temporal dimension. I3D not only reuses the architecture of 2D models, but also bootstraps the model weights from 2D pre-trained networks. I3D is able to learn spatial information at different scales and aggregate the results. It achieved the best performance (71.1\% top1 accuracy) on the Kinetics-400 dataset compared against previous works \cite{4}.

Action recognition models can extract spatio-temporal information efficiently, however, require huge computational power. To address this problem,Feichtenhofer et al. \cite{7} proposed SlowFast, an efficient network with two pathways, i.e., a slow pathway to learn static semantic information and a fast pathway focusing on learning temporal cues. Each pathway is specifically optimised for its task to make the network efficient. SlowFast 16 × 8, R101+NonLocal outperforms the previous state-of-the-art models, yielding 79.8\% and 81.8\% top-1 accuracy on the Kinetics-400 dataset and the Kinetics-600 dataset respectively \cite{7}. 

Feichtenhofer~\cite{8} introduced X3D networks to further improve the efficiency. X3D, a single pathway architecture also extended from a 2D image network. The most complex X3D model X3D-XL only has 11 million parameters, which is over 5 times less than SlowFast. Therefore, X3D is remarkably efficient. However, X3D-XL achieves better performance (81.9\% top1 accuracy) than SlowFast on the Kinetics-600 dataset and slightly lower performance (79.1\% top-1 accuracy) on the Kinetics-400 dataset with much lower computational cost \cite{8}.

Human drivers predict lane change intentions mainly use visual clues rather than  physical variables. However, existing works that utilize appearance features for lane change are surprisingly few. In \cite{39}, two appearance features, the state of brake indicators and the state of turn indicators are used for lane change recognition. In \cite{73}, the authors applied two action recognition methods, the spatio-temporal multiplier network and the disjoint two-stream convolutional network, to predict and classify lane change events on the PREVENTION dataset. Their method achieved 90.3\% classification accuracy. Izquierdo et al. \cite{25} considered lane change classification as an image recognition problem and utilised 2D CNNs. Their method achieved 86.9\% classification and 84.4\% prediction accuracy. Laimona et al. \cite{44} employed GoogleNet+LSTM to classify lane changes. This method yield 74.5\% classification accuracy. Although the methods stated above achieved decent performance, they all require additional information of the target change vehicle such as its contour and bounding box coordinates to pre-process input data. In contrast, our proposed method, RGB+3DN, does not require any extra information.

\section{Methods}

In this work, we propose an end-to-end framework involving two approaches for lane change recognition classification and prediction of surrounding vehicles in highway scenarios. Seven state-of-the-art 3D action recognition models are investigated including one I3D model, two SlowFast models and four X3D models for both our first and second approaches to address the lane change recognition problem. Different from all the existing methods, the first approach does not rely on extra information. Moreover, we further investigate the temporal and spatial attention region of 3D models by using a class activation map technique\cite{98CAM}, see section \ref{sec:experiment}. In this section, we firstly give the problem formulation of lane change recognition. The detail of our proposed framework are described in sections~\ref{RGB Video Data} and~\ref{sec:bbdata}. Figure~\ref{fig: architecture} illustrates the overall architecture of our method.

\subsection{Problem Formulation}
\label{Problem Formulation}
We consider lane change classification and prediction as an action recognition problem. 
Given an input video clip  $V \in \mathbb{R}^{F \times C \times H \times W}$, the goal of the task is to predict the possible lane change $\bm{y}\in{\{0,1,2\}}$ of this video where ${\{0,1,2\}}$ denotes lane keeping, left lane change and right lane change labels. Note that $F$, $C$, $H$, $W$ are the number of frames, channels, height and width  of an input video respectively and there is only one type of lane change in a video clip. The input video will then be sampled to a specific shape for different models and fed to the first convolution layer. As illustrates in Figure \ref{fig: architecture}, for I3D and SlowFast models, the features will then be sent to a max pooling layer and ResNet blocks. Whereas, for X3D models, the features will be directly sent to ResNet blocks. 
The features are convolved to $\mathbb{R}^{C \times T \times S^2}$, where C, T and $S^2$ denote channel, temporal and spatial dimension of the features. After residual convolution, the features are fed to one global average pooling layer and two convolution layers before fully connected layer. Softmax is finally applied for prediction, $ \hat{y} $. The loss function is simply the categorical cross-entropy:
\begin{equation}
L = -\frac{1}{N}\sum_{i=1}^{N} y_i \log(\hat{y}_i),
\label{equation loss}
\end{equation}
where $N$ is the number of classes.

For SlowFast networks, lateral connection is performed to fuse the temporal feature of the fast pathway and spatial feature of the slow pathway by concatenation after convolution and pooling layers, as shown in Figure \ref{fig: architecture}. Based on the methodology described above, we propose the following two approaches for utilising the same seven action recognition models but with different input data:
\begin{figure}[t]
    \centering
    \includegraphics[width=\textwidth]{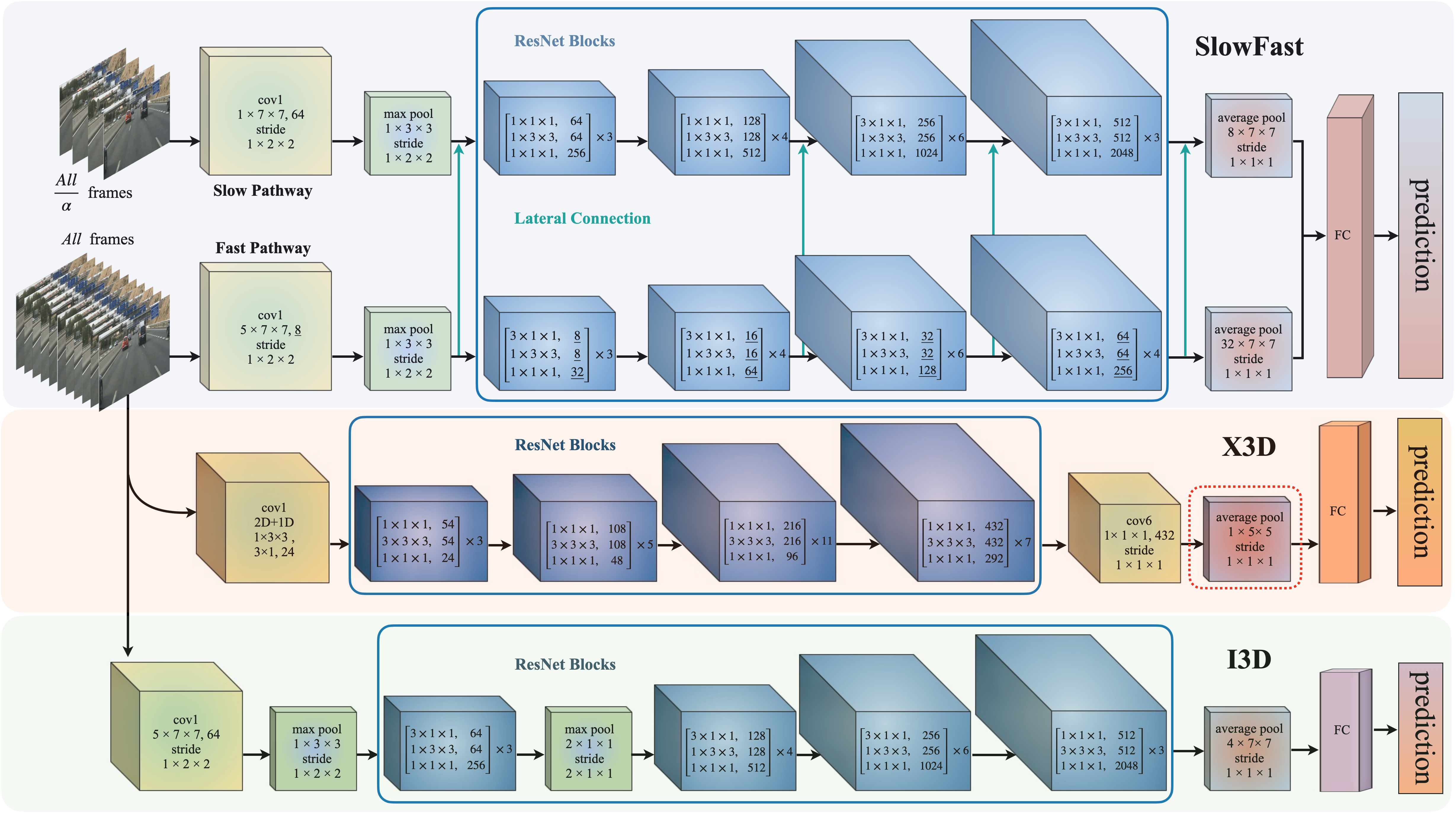}
    \caption{Architecture of models employed. This figure takes SlowFast$-$R50, X3D$-$S and I3D$-$R50 as example. I3D and X3D take all the frames of a video clip as input. The number of input frames of the fast pathway is \alpha  (\alpha = 8) times higher than the slow path way. The fast pathway has a ratio of \beta (\beta = $1/8$) channels (underlined) of the slow pathway. The red rectangle shows the temporal information extraction experiments conducted on the global average pooling layer of X3D$-$S.}
    \label{fig: architecture}
\end{figure}

\begin{figure}[t]
    \centering
    \includegraphics[width=0.75\textwidth]{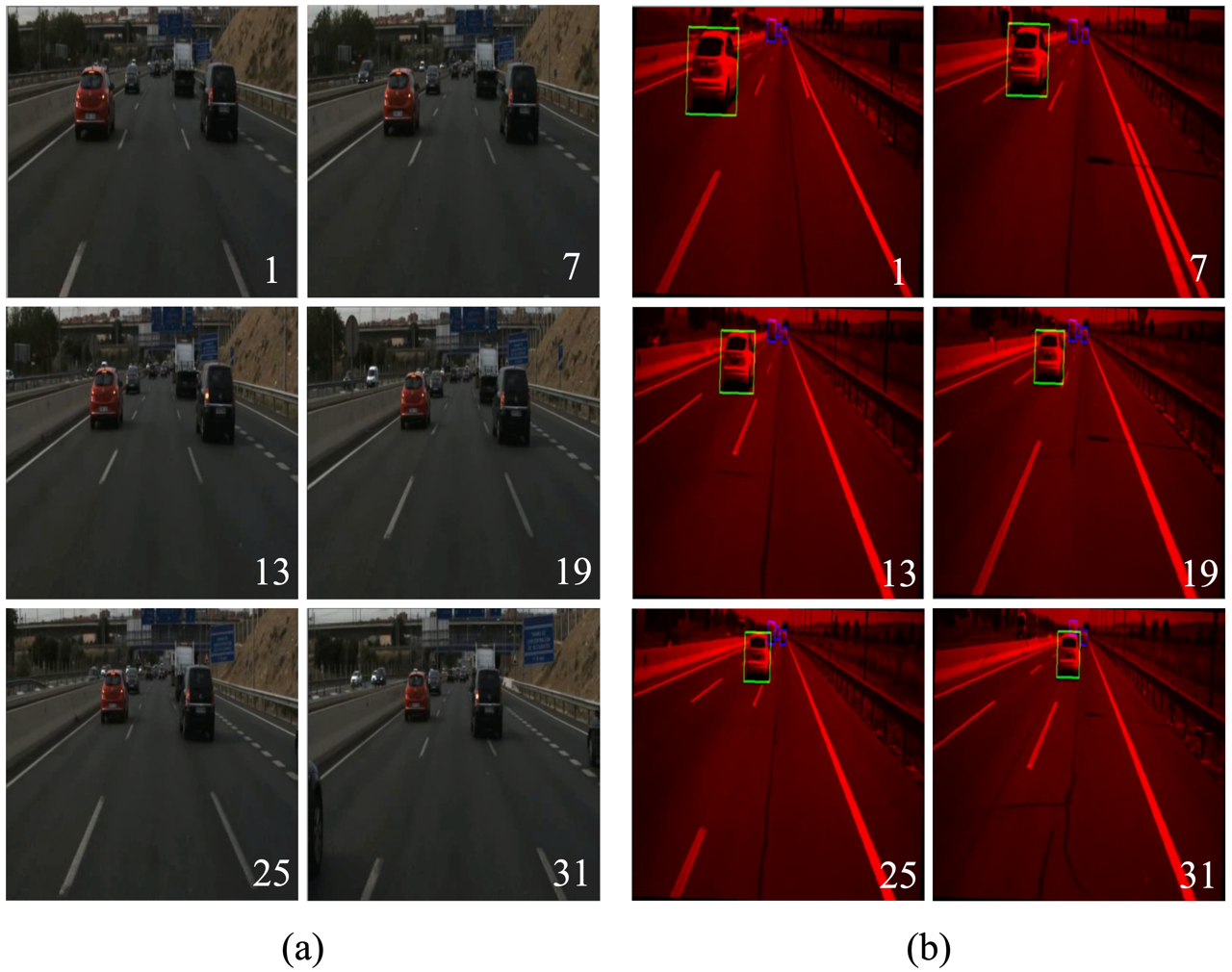}
    \caption{Input data visualisation (a) RGB video frame data of event; (b) video combined with bounding box data. Only the 1th, 7th, 13th, 19th, 25th and 31st frames are shown. A vehicle in the right lane in (a) and the left lane in (b) perform left and right lane change manoeuvres respectively. The frame data is resized to have aspect ratio of 1 ready for input to a classification CNN.}
    \label{fig: data_illust}
\end{figure}

\begin{description}
\item[RGB+3DN]: The first method utilises {\em only} the visual information collected by the front-facing cameras, which is the same kind of information and approach that human drivers would use to predict manoeuvres. We test this approach with seven 3D action recognition networks involving I3D networks, SlowFast networks, X3D networks and their variants. 
\item[RGB+BB+3DN]: The second method is designed over the first one. It uses the same 3D action recognition networks as the first method. Bounding box information is embedded to each frame of the RGB video data to improve classification and prediction accuracy. This method assumes that a separate vehicle prediction method has been used on the RGB input frames, prior to our lane change prediction. 
\end{description}

In order to assess the classification and prediction ability of the 3D CNNs, and following the practice given in \cite{73},  the concept of Observation Horizon (N) and Time To Event (TTE) is used to parameterize the temporal information fed to the models. As Figure \ref{fig:illst_tte} illustrates, the Observation Horizon is the time window before a target vehicle starts to perform a lane change event. Based on the investigation of Izquierdo et al. \cite{44}, the average length of an lane change event is typically 40 frames (4 seconds). TTE is defined as the time period from the specific time point to the time point of $f_1$. In other words, when TTE equals 20 (half average length of an lane change event, 2 seconds), the specific time point is $f_0$, which is when the target vehicle just starts performing lane change. The following parameters of input data with respect to Observation Horizon, and Time To Event are investigated in this work:
\begin{description}
\item[N40-TTE00]: Observation horizon: 40 frames (4 seconds), Time To Event: 0 frame (0 second), total data length of one sample is 60 frames (6 seconds). This dataset is designed to investigate the classification ability of the action recognition models, as the labeled lane change events have already occurred. 
\item[N40-TTE10]: Observation horizon: 40 frames (4 seconds), Time To Event: 10 frames (1 second), total data length of one sample is 50 frames (5 seconds). This dataset is designed to investigate the prediction ability (1 second ahead) of the action recognition models. 
\item[N40-TTE20]: Observation horizon: 40 frames (4 seconds), Time To Event: 20 frames (2 seconds), total data length of one sample is 40 frames (4 seconds). This dataset is designed to investigate the prediction ability (2 seconds ahead) of the action recognition models. 
\end{description}

\subsection{RGB+3DN: 3D Networks and RGB Video Data}
\label{RGB Video Data}

This section describes our first method, RGB+3DN.  Different from all the existing methods, it only requires the original RGB video as input to perform prediction.

We extract the samples of each class using the annotation provided by the PREVENTION dataset. The samples are initially centre-cropped from 1920 × 600 to 1600 × 600 pixels, then resized to $400$ × $400$ pixels in spatial resolution. In order to reduce the computational cost, the data is further downsampled from 10 FPS to 32 / 6 FPS in frame rate. Data augmentations are applied to expand the datasets. Figure \ref{fig: data_illust} (a) illustrates a sample of RGB video data.

\subsection{RGB+BB+3DN: 3D Networks and Video Combined with Bounding Box Data}
\label{sec:bbdata}

In this second approach, we employ the same seven CNNs as the first approach, however, the input video data used for this method is different and incorporates vehicle bounding box information. 
The method of processing video combined bounding box data is inspired by \cite{44}, where they employed 2D image classification models and still images combined with contours of vehicle motion histories.

To generate video combined bounding box data, we firstly use the same method of generating RGB video data to extract raw video clips of each class. Then the vehicle bounding boxes of vehicles in each frame are rendered into colour channels of each frame: the red channel is used to store the scene appearance as a gray scale image; the green channel to draw a bounding box of the target vehicle; and the blue channel to draw the bounding boxes of surrounding vehicles. These frames of each sample are then converted into video clips. Figure \ref{fig: data_illust} (b) illustrates a sample of video combined bounding box data. Note that the assumption here is that an estimated bounding boxes of vehicles in the scene and the location of the target vehicle  (the vehicle undertaking the lane-change) is available with the RGB input frames, perhaps by running a separate vehicle detection and tracking algorithm.

\section{Experiments}

\label{sec:experiment}
We evaluate and compare seven action recognition models on the PREVENTION dataset~\cite{izquierdo2019prevention} for lane change classification and prediction. This section describes the experiments in detail.

\subsection{Dataset}
The PREVENTION dataset consists of video footage, object bounding boxes, lane change events, trajectories, lane marking annotations, LiDAR and radar data. The dataset has 5 records. In each record, there is an annotation file named \textit{detections\_filtered.txt}, which contains the bounding box and contour information of all objects detected in the visual scene. The annotation information is stored in the form $[frame$, $ID$, $class$, $xi$, $yi$, $xf$, $yf$, $conf$, $n]$ where $class$ is the variable used to denote the types of the objects tracked, and $n$ is a sequence of $x$ and $y$ coordinates denote the contour of the tracked vehicle. The \textit{lane\_changes.txt} file contains the information with regard to the detected vehicle lane change events. The information is denoted in the format  $[ID$, $ID-m$, $LC-type$, $f0$, $f1$, $f2$, $blinker]$ where $ID-m$ is the unique ID of each vehicle performing the lane change, $LC-type$ denotes the type of each detected lane change event: 3 stands for left lane change and 4 stand for right lane change, $f0$ refers to the frame number at which lane change starts, $f1$ denotes the frame number at which the rear middle part of the vehicle is just between the lanes, $f2$ is the number of the end frame.

The data used for the first approach, RGB+3DN, and the second approach, RGB+BB+3DN, are RGB video data and video combined bounding box data respectively. Because of the limited  annotation available in the PREVENTION data, training and validation data used for the second approach is more limited than used for the first approach. Data augmentation methods such as random cropping, random rotation, colour jittering, horizontal flipping are performed to the all data. After data augmentation, the total number of each class of RGB video data and video combined bounding box data is 2420 and 432 respectively. For RGB video data, 1940 samples of each class are used for training and the rest 480 samples are used for validation. For video combined bounding box data, sample numbers of each class in the training set and the test set are 332 and 100.

\subsection{Evaluation Metrics}

We perform 4-fold cross validation to all methods. As a three-class classification problem, the accuracy was considered as the main metric to assess the performance of the networks employed in each scheme. The same as the evaluation metrics used in \cite{2}, the model accuracy was calculated by 
\begin{equation}
Acc = \frac{\textit{TP}}{\textit{Total No. of Samples}}
\label{equation 2.2}
\end{equation}
i.e. the number of true positive samples ($TP$) for the three classes divided by the total number of samples. The final top-1 accuracy is calculated by averaging the results of each fold.

%---------------------------------- Table 1&2 ----------------------------------------------

\begin{minipage}[t]{\textwidth}\hskip-0.5cm
\begin{minipage}[t]{0.54\textwidth} \small
\centering
     \makeatletter\def\@captype{table}\makeatother\caption{Top-1 classification and prediction accuracy of RGB+3DN method on the PREVENTION dataset. }
         \vspace{5 pt}
\label{table: RGB+3DN}
    \begin{tabular}{c|c|c|c|c} 
    \toprule  
    Model & GFLOPs & TTE-00 & TTE-01 & TTE-20 \\
    
    %\specialrule{\cmidrulewidth}{0pt}{0pt} 
    \hline
    X3D-XS        &\textbf{0.91} &82.78             &  75.69           &  64.10\\
    X3D-S         &2.96          &\textbf{84.79}    &  75.00           &  63.82\\
    X3D-M         &6.72          &82.36             &  69.72           &  63.19\\
    X3D-L         &26.64         &77.57             &  69.30           &  59.51\\
    I3D           &37.53         &83.05             &  73.75           &  63.54\\
    SF,R50        &65.71         &81.04             &  \textbf{76.67}  &  \textbf{65.00}\\
    SF,R101       &127.20        &71.80             &  74.79           &  62.08\\
    \bottomrule 

    \end{tabular}
  \end{minipage}
  \begin{minipage}[t]{0.46\textwidth} \small
   \centering
        \makeatletter\def\@captype{table}\makeatother\caption{Top-1 classification and prediction accuracy of RGB+BBS+3DN method on the PREVENTION dataset.}
        \label{table: BBS+3DN}
        \vspace{5 pt}
        \begin{tabular}{c|c|c|c} 
        \toprule  
        Model&TTE-00& TTE-01& TTE-20 \\
        
        %\specialrule{\cmidrulewidth}{0pt}{0pt} 
        \hline
        X3D-XS        &98.33             &  \textbf{99.17}  &  \textbf{98.86}\\
        X3D-S         &\textbf{99.33}    &  98.33           &  97.82\\
        X3D-M         &98.17             &  98.67           &  97.19\\
        X3D-L         &98.43             &  98.34           &  97.51\\
        I3D           &98.67             &  98.84           &  96.54\\
        SF,R50        &98.50             &  98.50           &  98.67\\
        SF,R101       &97.33             &  97.67           &  96.33\\
        \bottomrule

      \end{tabular}
   \end{minipage}
\end{minipage}

\subsection{Lane Change Classification and Prediction with RGB+3DN}

\subsubsection{Lane Change Classification}

Table \ref{table: RGB+3DN} presents the top-1 accuracy of lane change classification and prediction on RGB video data. Among all the models, SlowFast-R101 (with 127.2 GFLOPs) is the most complex model. It outperforms all the other models on the Kinetics-400 dataset \cite{7}. However, SlowFast-R101 networks only yield 71.80\% top-1 accuracy on RGB video data, which is the lowest. Whereas, the relatively lightweight SlowFast-R50 networks, with nearly half GFLOPs, achieve 81.04\% top-1 accuracy. Surprisingly, although all the four X3D models are much more lightweight than the SlowFast models, they outperform all the SlowFast models. The second most lightweight model X3D-S achieves the best performance (84.79\%  top-1 accuracy). The I3D model is the second best model. It yields 83.05\% top-1 accuracy, which is only 1.74\% lower than the best model.

An interesting finding can be observed from the results described above, i.e., in general, for lane change lane change classification, the lighter the model, the better the accuracy is.
This could be explained because the capacity of the models employed is large, but the dataset does not have sufficient variation. 

\subsubsection{Lane Change Prediction}
Models can anticipate 1 second and 2 seconds ahead on TTE-10 and TTE-20 data respectively. The prediction accuracy of all the models decrease more significantly than their classification accuracy. This can be explained as TTE-10 and TTE-20 data provide less information than TTE-00 data. As can be seen in Table \ref{table: RGB+3DN}, the best model on both TTE-10 and TE-20 data is the SlowFast networks, which yield top-1 accuracy of 76.67\% and 65\% for 1 and 2 seconds anticipation respectively. On the prediction experiments, a similar finding as on the classification experiments is observed, i.e. higher model complexity leads to lower prediction accuracy. For instance, SlowFast-R50 is more lightweight than SlowFast-R101. However, SlowFast-R50  outperforms SlowFast-R101 by nearly 2\%. This finding also applies to the all X3D models.

\begin{figure}[t]
    \centering
    \includegraphics[width= 0.8\textwidth]{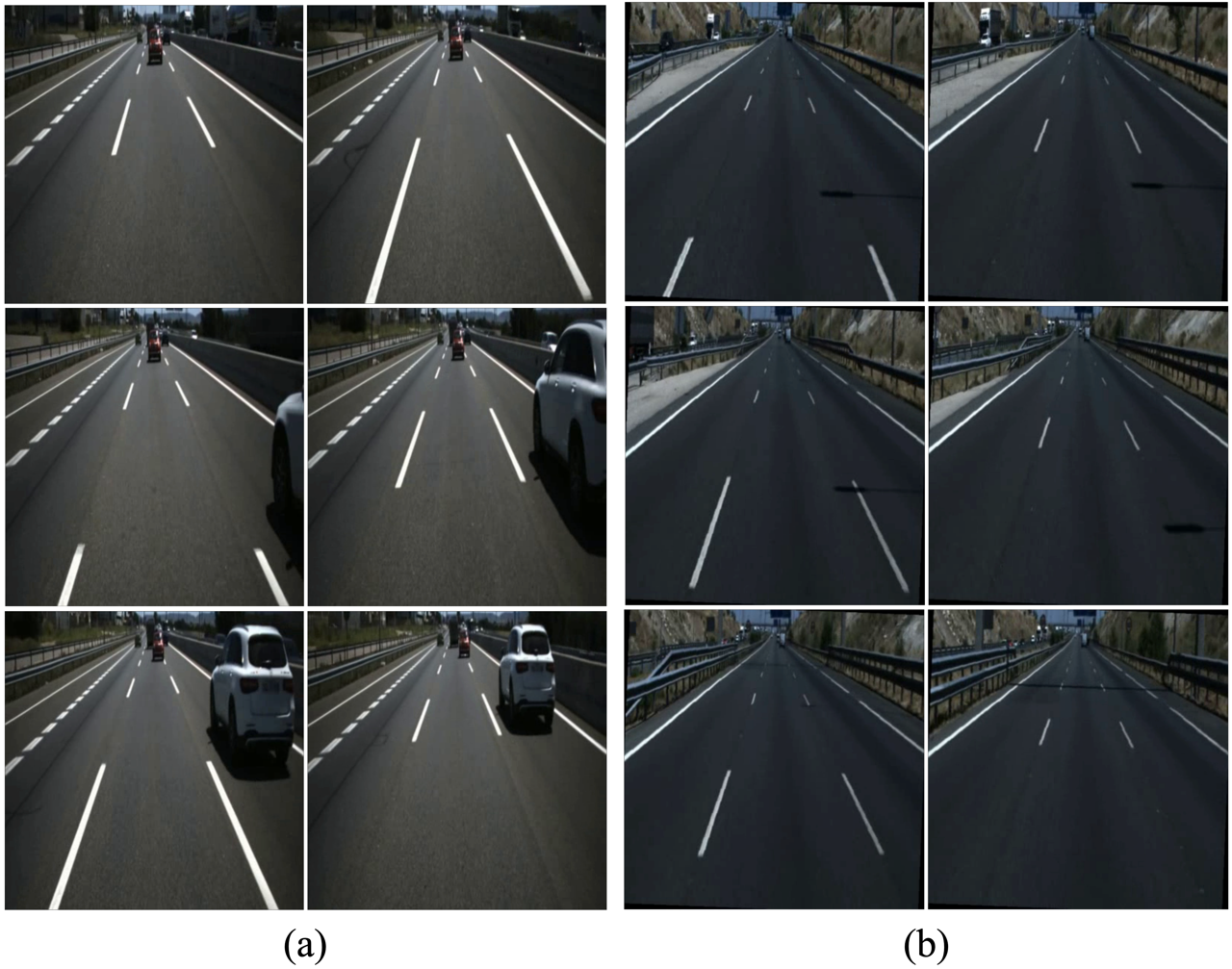}
    \caption{TP and FP examples: (a) TP example, the vehicle in the middle performing right lane change is correctly classified. (b) FP example, the vehicle in the right lane performing left lane change is classified as lane keeping incorrectly due to small target.}
    \label{fig:goodbad_example}
\end{figure}
By analysing classification and prediction confusion metrics of the RGB+3DN method, we observe that, for lane change classification, the accuracy of right lane change class is higher than left lane change. This can be explained as many of the Left Lane Change events take place in areas far from the ego vehicle and the target vehicles are relatively small, as can be seen in Figure \ref{fig:goodbad_example} (b). Whereas, right lane changes normally take place in front of the ego vehicle, which is close to the cameras, as Figure \ref{fig:goodbad_example} (a) illustrates: clearer footage results in better image representations and better recognition accuracy. Lane keeping is always the best predicted class for TTE-00, TTE-10 and TTE-20 data, as its scene is relatively simple and easy to predict.
For lane change prediction, the accuracy of lane keeping class is significantly higher than the other two classes. This may be due to there being fewer frames given to these events, some lane and right lane changes also resemble lane keeping events.
Therefore, left lane and right lane changes can be sometimes miss-predicted as lane keeping.

\subsection{Lane Change Classification and Prediction with RGB+BB+3DN}

Table \ref{table: BBS+3DN} illustrates the classification and prediction results on video combined with bounding box data.  As can be observed, regardless of the classification or prediction results, the performance of each method does not vary much. The best accuracy is only 3.00\% higher than the lowest one, which is very different from the experiments of the RGB+3DN method on the RGB video data. The X3D-L and SlowFast-R101 are always the two lowest performing models. Whereas, the best performing models are always from the X3D family. 

The temporal anticipation accuracy is not affected much in any of the models, although there is still a 1\% to 1.5\% drop in performance between TTE-00 and TTE-20 in all cases. The prediction accuracy however is considerably enhanced by knowing the location of vehicles by their bounding boxes giving much higher performance than with RGB information alone. 

% This shows the boost that the temporal information affords in solving this problem. 

%---------------------------------- Table 3 ----------------------------------------------

\begin{table}[!t]
\caption{Comparison of different methods on PREVENTION dataset.}
\label{table: compare to exsiting methods}
\centering
\begin{tabular}{c|c|c|c|c}
\toprule  
Method& Extra Information& TTE-00& TTE-01& TTE-20 \\
%\specialrule{\cmidrulewidth}{0pt}{0pt} 
\hline
GoogleNet + LSTM \cite{44}  & ROI + Contour&  74.54         &   -           &  -\\
2D Based  \cite{25}          & Contour      &  86.90         &  84.50        &  -\\
Two Stream Based \cite{73}   & ROI          &  90.30         &  85.69        &  91.94\\
VIT  \cite{konakallacnn}     & Non          &  81.23         &   -           &  -\\
\hline
Ours (RGB + 3DN)             &\textbf{Non}  & 84.79          &  76.67        &  65.00\\
Our (RGB + BB+ 3DN)         & BBS          & \textbf{99.33} &\textbf{99.17} &\textbf{98.86}\\

\bottomrule 
\end{tabular}
\end{table}

%--------------------------------------------------------------------------------

\subsection{Comparison to previous Methods}
As Table \ref{table: compare to exsiting methods} illustrates,  Simonyan et al.'s two stream based method obtains better accuracy than our RGB+3DN method, however, their method requires bounding box coordinates of the target vehicle for Region of Interest (ROI) cropping. Furthermore, their validation data of each class is highly unbalanced. Although our RGB+3DN only uses the original video data collected by cameras and does not require additional information, we still outperform some methods, e.g., GoogleNet + LSTM and VIT. 

The method for processing the data of our RGB+BB+3DN method is inspired by Izquierdo et al.’s work \cite{44}. In \cite{44}, their best performing model, GoogleNet + LSTM yields 74.4\% for lane change classification. Because 3D models can better extract spatio-temporal features than 2D CNNs, our RGB+BB+3DN method achieves top-1 classification accuracy of 99.33\%, which is significantly higher. With extra bounding box information, our RGB+BB+3DN method out-performs all the methods compared.

\begin{figure}[t]

    \centering
    \includegraphics[width= \textwidth]{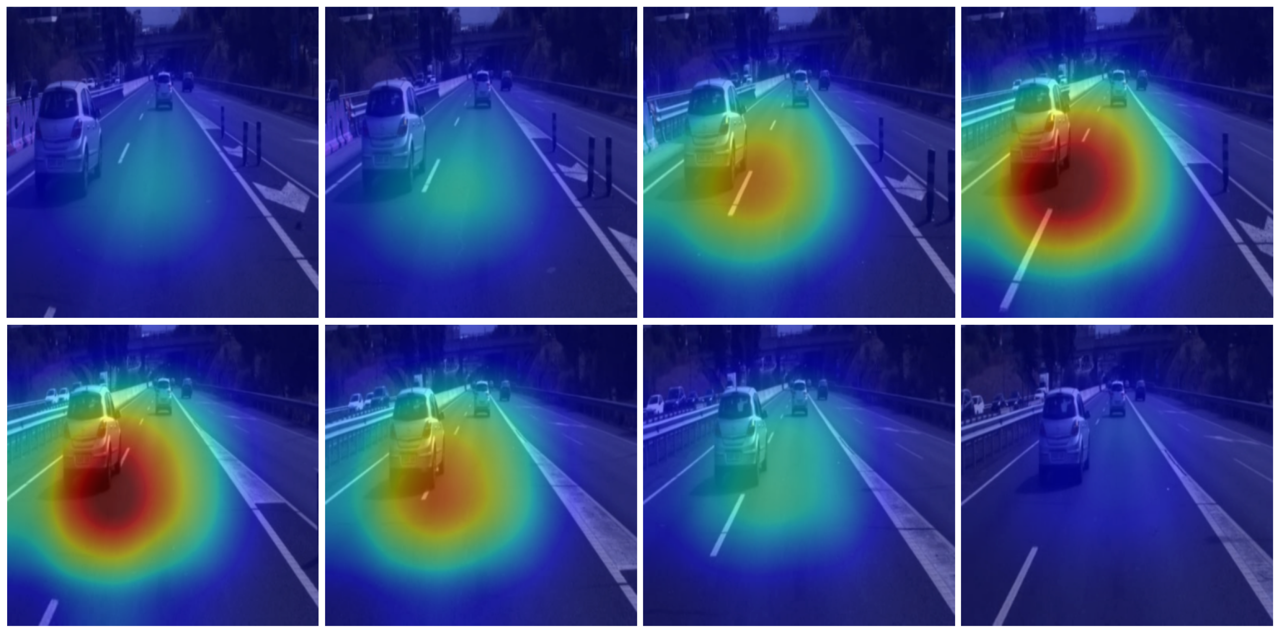}
    \caption{Class activation maps. Only the 25th to 32nd frames of the input video are shown. The model mainly focuses on the frames where lane change happens, as well as the edge of the target vehicle and lane marking which it is about to cross. }
    \label{fig:cam}
\end{figure}

\subsection{Class Activation Maps}
To investigate whether our models recognize the motion clues and learn spatio-temporal information, we generated class activation maps on X3D-S and RGB video data. The features of the last convolution layer were used to calculate the CAMs. The calculated scores were normalised across all patches and frames before visualisation. Experimental results show that action recognition models can efficiently extract both spatial and temporal information. As Figure \ref{fig:cam} illustrates, in the spatial domain, the model focuses on the edge of the target vehicle and lane marking which the target vehicle is about to cross. Whereas, in the temporal domain, the X3D-S only focuses on those frames that the lane change event happens, specifically, frames 27 to 30.

\subsection{Optimizing Temporal Information Extraction}
\begin{figure}[t]
    \centering
    \includegraphics[width=0.75\textwidth]{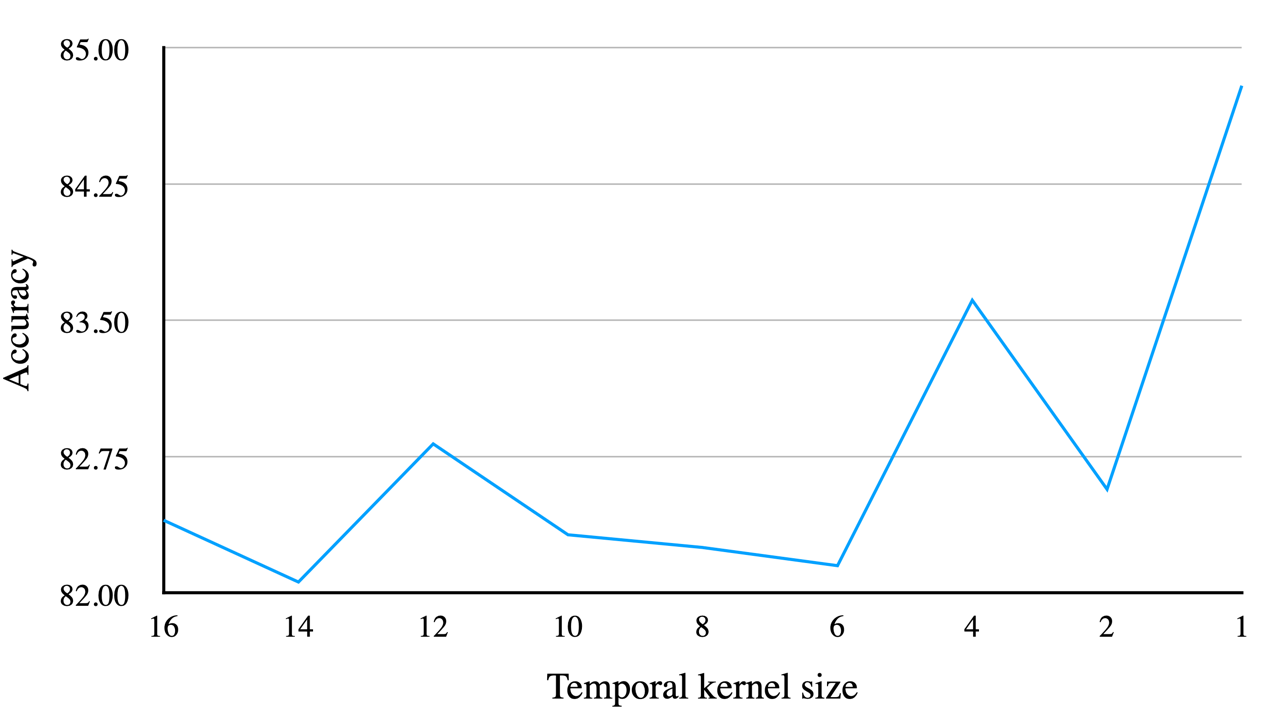}
    \caption{Experiments on temporal kernel size. Smaller kernel size results in better accuracy. }
    \label{fig:kernel}
\end{figure}

The 3D CNNs we employ in this work are initially designed for recognising general human behaviours and trained on human behaviours datasets such as Kinetics-400 and Kinetics-600. These datasets are formed by video clips with relatively high frame rates (25 fps) \cite{4}. Therefore, in order to efficiently extract motion clues,  the temporal kernel size of the global layer (as shown in the red rectangle of Figure \ref{fig: architecture}) of X3D is originally designed as 16. Whereas, the frame rate of our data is much less, 32/6. Moreover, while we inspected class activation maps, an interesting finding was revealed, i.e., the model only pays attention to the 3rd to 5th frames where the target vehicle approaches the lane marking and overlay it, as shown in Figure \ref{fig:cam}. Therefore, we postulated that a larger temporal kernel size could potentially introduce noisy information, and decreasing the temporal dimension size of the kernels might improve the model performance. This seems to be the case. As Figure \ref{fig:kernel} shows, as we decrease the temporal kernel size from 16 to 1, the model classification accuracy increases by 2.39\% from 82.40\% to 84.79\%.

\section{Conclusions}
Two approaches involving seven 3D action recognition networks are adopted to classify and anticipate lane change events on the PREVENTION dataset. For our RGB+3DN method, the lane change recognition problem is formulated as an action recognition task only utilising visual information collected by cameras. The best performing model, X3D-S achieves state-of-the-art top-1 accuracy (84.79\%) for lane change classification using the original RGB video data of the PREVENTION dataset. Our RGB+BB+3DN method achieves significant accuracy improvement (TTE-20 98.86\%) by taking the advantage of 3D CNNs and additional bounding box information. Furthermore, we generated class activation maps to investigate the spatial and temporal attention region of 3D CNNs. These CAMs demonstrated that action recognition models are able to extract lane change motions efficiently. We proposed a way to better extract relevant motion clues by decreasing the dimension of the temporal kernel size.

As further work, more data needs to be added for training and evaluation to expand the diversity of the dataset and prevent overfitting. Introducing vehicle and lane detectors to the 3D networks to further exploit appearance clues of the data and improve model performance is one potential strategy. Although still challenging, in this way, the model might achieve the performance of RGB+BB+3DN method of this work without separate bounding box/target vehicle information being required.

%%%%%%%%%%%%%%%%%%%%%%%%%%%%%%%%%%%%%%%%%%%%%%%%%%%%%%%%%%%%%%%%%%%%%%%%%%%%%%%%%%%%%%%%%%%%%%%

% \clearpage\mbox{}Page \thepage\ of the manuscript.
% \clearpage\mbox{}Page \thepage\ of the manuscript.

% This is the last page of the manuscript.
% Now we have reached the maximum size of the ECCV 2022 submission (excluding references).
% References should start immediately after the main text, but can continue on p.15 if needed.

% \clearpage
\bibliographystyle{splncs04}
\bibliography{egbib}

\end{document}